\documentclass{article}

\usepackage[utf8]{inputenc}
\usepackage[all]{xypic}
\usepackage{amssymb}
\usepackage{amsmath}
\usepackage{graphicx}
\usepackage{booktabs} 
\usepackage{hyperref}
\usepackage{caption}
\usepackage{subcaption}
\usepackage{enumitem}
\usepackage{amsmath}
\usepackage{array}
\usepackage{url}            
\usepackage{microtype}
\usepackage{booktabs}
\usepackage{amsfonts}       
\usepackage{multirow}
\usepackage{array}

\usepackage[accepted]{icml2020}

\newcommand{\PreserveBackslash}[1]{\let\temp=\\#1\let\\=\temp}
\newcolumntype{C}[1]{>{\PreserveBackslash\centering}p{#1}}
\newcolumntype{R}[1]{>{\PreserveBackslash\raggedleft}p{#1}}
\newcolumntype{L}[1]{>{\PreserveBackslash\raggedright}p{#1}}
\newcolumntype{M}[1]{>{\centering\arraybackslash}m{#1}}

\DeclareMathOperator*{\argmax}{arg\,max}

\icmltitlerunning{Explaining Deep Neural Networks using Unsupervised Clustering}

\begin{document}

\twocolumn[
\icmltitle{Explaining Deep Neural Networks using Unsupervised Clustering}

\begin{icmlauthorlist}
\icmlauthor{Yu-Han Liu}{to}
\icmlauthor{Sercan \"{O}. Ar{\i}k}{to}
\end{icmlauthorlist}

\icmlaffiliation{to}{Google Cloud, Sunnyvale, CA}

\icmlcorrespondingauthor{Yu-Han Liu}{yuhanliu@google.com}

\icmlkeywords{Interpretability, Concept-based Explainability, Sample-based Explainability}

\vskip 0.3in
]


\begin{abstract}
We propose a novel method to explain trained deep neural networks (DNNs), by distilling them into surrogate models using unsupervised clustering. Our method can be applied flexibly to any subset of layers of a DNN architecture and can incorporate low-level and high-level information. On image datasets given pre-trained DNNs, we demonstrate the strength of our method in finding similar training samples, and shedding light on the concepts the DNNs base their decisions on. Via user studies, we show that our model can improve the user trust in model's prediction.
\end{abstract}

\vspace{-0.7cm}
\section{Introduction}
\vspace{-0.2cm}

Artificial Intelligence (AI) is the core of information processing in many applications, fueled by the recent progress in deep neural networks (DNNs). One major bottleneck for wider spread adoption is `explainability'  \cite{rudin_explainableAI} \cite{alej2019explainable}. In their conventional form, DNNs are `black box' models, where decision making is controlled by complex non-linear interactions between many parameters.

To explain DNNs, numerous ideas have been explored, including activation visualizations \citep{Bolei2018, visualizeHF, visualizeconv2}, attribution methods \citep{lundberg2017unified, deeplift}, concept activations \citep{ghorbani2019automatic, kim2017interpretability}, and distillation methods \citep{frosst2017distilling}. 
One understudied aspect is explaining what a DNN learns at different layers, as the information is being processed from raw input features to output labels. Since early days of DNNs, it is well-known that earlier layers focus more on low-level information, whereas later layers focus more on high-level information \cite{cnn_filters1}. Exposing this to humans, in a way that they can easily understand and get actionable insights, is a challenging problem.
While relating one example to another and making classification decisions, humans pay attention to low-level details, as well as the high-level concepts \citep{kahneman2011thinking}. We hypothesize that both low- and high-level information should be useful for explainability purposes -- while low-level features can directly show the immediate similarity patterns (attributed to fast thinking), high-level features can show high level conceptual similarities (attributed to slow thinking). When an AI system is considered, in a similar vein, to explore its rationale to humans, we argue that interpretation of information at different layers should be important. We need a systematic framework that allows explainability from low-level and high-level reasoning perspectives.

In this paper, we propose a novel explainability method by shedding a light on how a `black-box' a DNN processes similar input samples at different layers. Our method is based on the fundamental intuition -- \emph{the processed representations at different layers can be systematically categorized without losing too much salient information}. We propose an unsupervised clustering-based model distillation approach. We show that our surrogate model can obtain high fidelity with the baseline black-box DNN, while being explainable by design. To demonstrate explanation capabilities of the proposed surrogate model, we focus on sample-based and concept-based explanation scenarios. We show that our method can be very powerful at (i) providing insights on the low- and high-level patterns learned by the model, (ii) identifying the most relevant samples to support a decision, according to human judgement of sample similarity, (iii) clustering the sub-concepts within the labels.

\vspace{-0.3cm}
\section{Related Work} 
\vspace{-0.4cm}

\noindent\newline{\textbf{Distilling into interpretable models:}}
Distilling a pre-trained black-box model into an inherently-interpretable model is one of the common methods for explainability. In \citealt{frosst2017distilling}, DNNs are distilled into soft decision trees by matching the predicted distributions.
\citealt{tan2018learning} applies distillation to get global additive explanations.
\citealt{lime} proposes distillation locally for each instance, which is improved by \citealt{yoon2019rllim} by utilizing reinforcement learning to optimize fidelity of the locally-interpretable model.
Similar to conventional distillation approaches, we use a surrogate model, yet our approach of extracting it is fundamentally different.
\noindent\newline{\textbf{Sample-based explainability:}}
Visualizing the most `similar' samples is one common approach to explain DNNs. While some approaches like \citealt{li_protototypes} and \citealt{attention_prototypical} modify the design of DNNs, there are also approaches that bring sample-based explainability to pre-trained DNNs, like \citealt{kim2016examples}, \citealt{influence_functions} and \citealt{yeh2018representer}, similar to our approach. But none of these systematically analyze how low-level vs. high-level information is utilized for explainability. Our approach goes beyond heuristic assumptions on similarity using some distance metric in some activation space. Indeed, we propose a distillation method that is aimed to mimic the model while ``categorizing" the information at different layers.
\noindent\newline{\textbf{Concept-based explainability:}}
DNNs build conceptual intelligence on low-level features. Different approaches are proposed to uncover the concepts learned by pre-trained DNNs. \citealt{kim2017interpretability} quantifies the relationship to concepts are provided by humans whereas in \citealt{ghorbani2019automatic} concepts are automatically learned by clustering activations of input segments by a helper model.
\citealt{yeh2019completenessaware} proposes a concept discovery method with `completeness', by additionally encouraging them to be interpretable. These assume that in the activation space, some certain directions correspond to concepts. But neither they go beyond this assumption for concept discovery, nor they provide a systematic framework to consider different layers.
\noindent\newline{\textbf{Unsupervised clustering of representations:}}
DeepCluster \citep{deep_cluster1} applies joint training for the parameters of a DNN and the cluster assignments of the resulting features. By investigating the top activated images for a given filter, DeepCluster can show similar-looking images. DeeperCluster \citep{deep_cluster2} shows improvements over DeepCluster using self-supervised learning. While such methods modify the baseline DNNs, our method is a post-hoc explainability method. Besides, unlike them, our method considers all filters in a layer, by extracting the joint information from them with the encoder and decoder.

\vspace{-0.4cm}
\section{Learning a surrogate model}
\vspace{-0.2cm}

\subsection{Framework}

\begin{figure*}[!htbp]
\centering
\includegraphics[width=0.65\linewidth]{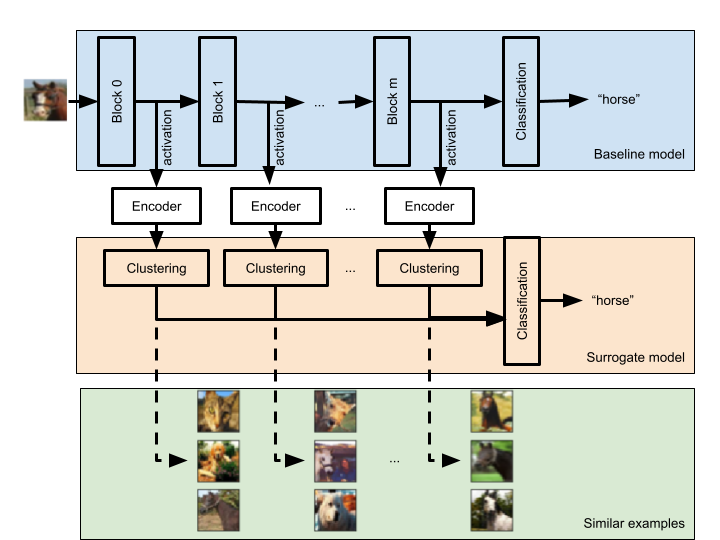}
\vspace{-0.4cm}
\caption{Our method is based on extracting information from the intermediate activations at \emph{multiple layers} of a learned baseline model, and then systematically grouping them together to relate the groups to the output decisions. Grouping the learned representations is based on unsupervised clustering, due to the lack of labels for the task. The dashed arrows represent highest-ranked training examples in the embedding space according to Euclidean distance. This way, each selected activation provides a notion of similarity which can be aggregated with controllable weights.}
\vspace{-0.2cm}
\label{diagram:design}
\vspace{-0.2cm}
\end{figure*}

Consider a baseline ``black-box'' trained on $\{(x_i, y_i)\}$, where $x_i$ are input examples and $y_i\in C=\{c_1, c_2, \ldots, c_n\}$ are the class labels. For each input $x$, let $f(x)$ denote the predicted posterior distribution $p(c|x)$ such that the predicted class is given by $\argmax f(x)$.
We assume that the model is architecturally `deep', composed of $L$ layers:
\vspace{-0.15cm}
\begin{equation}
x\longrightarrow a^1 \longrightarrow a^2 \longrightarrow ... \longrightarrow a^L \longrightarrow c,
\vspace{-0.15cm}
\end{equation}
where $a^j$ is the activation of the $j^{th}$ layer.

Fig. \ref{diagram:design} overviews our approach. Our aim is to learn ``fine-grained latent clusters'' $Z^j=\{z^j_1, z^j_2, \ldots, z^j_{\ell_j}\}$ for each layer $j$, to use them in systematic investigation of similarities. The latent clusters are trained to extract relevant information which the baseline learned to recognize during its training at the corresponding layer.
Ideally, a simple mapping from the latent cluster assignments would suffice to yield comparable performance with the baseline model. Yet, by grouping the high-dimensional intermediate activations into a small number of clusters and assigning the centroids to represent information, some content from each layer would be lost, but how much? We show that indeed a simple interpretable model based on empirical observations can map the latent clusters to the output decision with a very small sacrifice in the overall performance. 

The fine-grained latent clusters would represent concepts that the baseline model has learned beyond the information granularity of the labels (for which the baseline DNN was trained) or readily recognizable to humans. E.g., the class of horse images might get split into (possibly overlapping) sub-classes of those showing the entire horse; those showing only the head and neck; those with a rider, etc. Such conceptual information is generally captured by higher-level layers. On the other hand, there are easily-recognizable patterns, such as whether an image includes dirt or grass, that are readily available from low-level layers. 

For any given layer $j$, we propose a surrogate model $f_j = h_j \circ g_j$, composed of (i) a function $g_j$ that maps each example to a predicted posterior distribution of latent clusters $p(z^j|x)$ , and (ii) a function $h_j$ that maps assignments to the latent clusters, to predicted classes.
That is, we consider the underlying probabilistic model 
\vspace{-0.15cm}
\begin{equation}
x\longrightarrow \{z^j_1, z^j_2, \ldots\} \longrightarrow c
\vspace{-0.15cm}
\end{equation}
for the baseline DNN's reasoning for $x \longrightarrow c$, using information learned up to layer $j$, reducing the problem of explainability to the comparison of the latent clusters $Z^j$ with concepts which humans can recognize.  
The surrogate models of different layers can be averaged, potentially with different weights, to define a full surrogate model for the baseline DNN. Such weights would allow the user to put more emphasis in low-level or high-level concepts.
We propose the following desiderata for the full surrogate model:
\begin{itemize}[noitemsep, nolistsep, leftmargin=*]
    \item \textbf{Fidelity}: It should behave similarly to the baseline model and thus retain similar accuracy.\footnote{We quantify fidelity as the accuracy score for the activation clustering model to predict the baseline model's predictions.}
    \item \textbf{Usability}: It should not be more complex than the baseline model. Ideally, a simple operation should map the inputs to the explanations, as well as predictions. 
    \item \textbf{Customizability}: It should be customizable in an intuitive way, to balance the contribution of low- and high-level information as desired.
    \item \textbf{Interpretability}: It should be interpretable to humans, to explain what the DNN has learned locally and globally.
\end{itemize}

\vspace{-0.3cm}
\subsection{Architecture}
\vspace{-0.2cm}



We pick a subset of intermediate activations of the model's hidden layers (that are potentially high dimensional). Ideally, we would like to consider all activations, but that is computationally prohibitive. On the other hand, we expect that the difference of extra learned information between adjacent layers is small, so we should allow sufficiently large subgraph in between the selected activations. We then train an unsupervised clustering algorithm for every selected activation, using training examples' activations as input.

The clustering algorithm we adapt is the Deep Embedded Clustering (DEC) \cite{xie2015unsupervised}. DEC first trains an autoencoder whose encoder maps into an embedding vector space $V$, then fine-tune the encoder while learning cluster centroids in $V$ with a loss function based on the Euclidean distance. DEC outputs a discrete probability distribution over the set of clusters based on the $t$-distribution. To avoid scaling issues when we compare the embeddings of different layers, we normalize the embeddings so that they are on a hypersphere.\footnote{The radius of the hypersphere is set to be $8.0$ for all layers.} To encourage sparsity, we use a large normality parameter\footnote{We use $\alpha=100.0$ as opposed to the default value of $1.0$.} to reduce the $t$-distribution's variance.

We train a clustering model for each activation $a^j$, which consists of an encoder that maps the activations into an embedding space $V^j$ along with a probability assignment map into the set $Z^j=\{z^j_1, z^j_2, \ldots, z^j_{ \ell_j}\}$ of clusters.  These define a mapping from input examples to clusters.
To define the surrogate model for layer $j$, we need to define also $h_j$.  For this we use the empirical posteriors $p(y|z^j)$ observed from the training data. In other words, the $j$-th surrogate model's prediction is given by 
\vspace{-0.2cm}
\begin{equation}
p(y|x) \sim \sum\nolimits_{z^j_k\in Z^j}p(y|z^j=z^j_k)p(z^j|x).
\vspace{-0.15cm}
\end{equation}
The full surrogate model is a weighted average of the layer-wise surrogate models $h_j\circ g_j$. In our experiments, we observe that using equal weights for the layers is a reasonable choice in most cases, and we do so unless explicitly stated otherwise. In general, the user can conveniently adjust the weights based on whether lower- or higher-level information is more relevant for the specific dataset and task (e.g. by probing the explanations on a validation set).

\vspace{-0.3cm}
\section{Explanations from surrogate model}
\vspace{-0.2cm}

The surrogate model obtained via clustering of activations of different layers, is used to explain the baseline DNN model.
\noindent\newline{\textbf{Sample-based:}}
Given a test example, we find similar training examples based on the distances in the embedding spaces. 
For some tasks, low-level similarities may be more important to the human users (such as in medical anomaly detection), while for some other tasks, high-level similarities may be important (such as in scene classification).
For sample-based explainability, activations from different layers may be used and our framework enables it via weights. 
Note that our proposed similarity reflects the similarity \emph{according to the baseline/surrogate model} and humans need not agree.\footnote{Some works, e.g. \cite{ghorbani2019automatic}) use a different notion of similarity, using the features extracted by another deep learning model trained on a possibly unrelated but larger scale dataset such as ImageNet.}
For a sample-based explainability method, the quality should be measured by how much the similar examples support the model's decision - e.g., would humans become more inclined to accept the model's prediction (which need not be correct) upon seeing the similar examples?
\noindent\newline{\textbf{Concept-based:}}
Our model identifies a number of latent clusters of training examples for each selected layer.  The clusters are represented by the cluster centroids in the corresponding embedding vector spaces. We use these cluster centroids to extract the \emph{concepts} that reflect what the baseline model learns during training. We observe these concepts are often human-recognizable, qualitatively.

\vspace{-0.3cm}
\section{Experiments}
\vspace{-0.2cm}

\subsection{CIFAR10 dataset}
\vspace{-0.4cm}

\begin{table}[!htbp]
 \caption{Activations used for the surrogate model.\footnote{ The activation names refer to the layers in a \href{https://github.com/tensorflow/models/blob/9b98e3db132f64f28ffd1697c3f2cf79cfdaf37e/official/vision/image_classification/resnet_cifar_model.py}{reference ResNet implementation} which are are the outputs of four ResNet stages (the first stage being a single convolutional layer applied to the input image).}}
 \centering
 \begin{tabular}{|c|c|c|c|} 
 \hline
 Name & Shape & $\dim(V^j)$ & $\ell_j$ \\ [0.5ex] 
 \hline\hline
 $\text{activation}$ & 32, 32, 16 & 20 & 15  \\ 
 \hline
 $\text{activation\_18}$ & 32, 32, 16 & 20 & 15  \\
 \hline
 $\text{activation\_36}$ & 16, 16, 32 & 20 & 15  \\
 \hline
 $\text{activation\_54}$ & 8, 8, 64 & 20 & 15  \\
 \hline
\end{tabular}
\label{table:activations}
 \vspace{-.4cm}
\end{table}

\vspace{-0.3cm}

\begin{table}[!htbp]
\caption{Accuracy and fidelity on CIFAR10 test data.}
\centering
\begin{tabular}{|c|c|c|} 
 \hline
  Model & Accuracy & Fidelity \\ \hline 
  Baseline DNN & 0.8690 & 1.0 \\ \hline
  Surrogate & 0.8448 & 0.923 \\ \hline
\end{tabular}
\label{table:acc_fid}
 \vspace{-.4cm}
\end{table}

 \vspace{-.4cm}
\noindent\newline{\textbf{Setup:}}
On CIFAR-10 dataset, we use a pre-trained ResNet-56 model. We select activations and train DEC models with parameters in Table \ref{table:activations}; recall that $V^j$ is the embedding space for activation $a^j$, and $\ell_j$ is the number of clusters. Table \ref{table:acc_fid} shows that on test data, the simple surrogate model achieves similar performance with the baseline DNN.\footnote{On the training data, the surrogate model achieves $1.0$ accuracy and $1.0$ fidelity.}

\begin{figure}[!htbp]
\centering
\begin{subfigure}{0.49\textwidth}
  \centering
  \includegraphics[width=\linewidth]{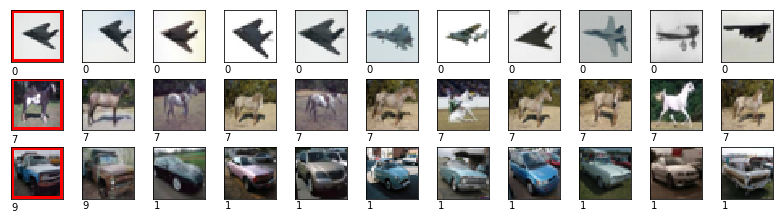}
   \vspace{-.6cm}
\caption{Proposed method with equal weights.}
\end{subfigure}
\begin{subfigure}{0.49\textwidth}
  \centering
  \includegraphics[width=\linewidth]{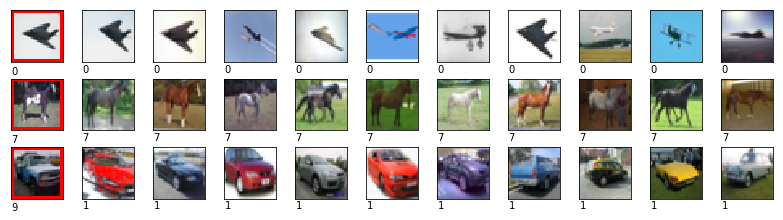}
   \vspace{-.6cm}
\caption{Nearest neighbors of extracted features.}
\end{subfigure}
  \vspace{-0.3cm}
\caption{Comparison of similar images between the proposed approach and applying Euclidean distances of the extracted features (namely, the activation that feeds into the final fully connected decision layer).  The left most column has the test images with red border, and ten most similar images are shown.  For each image, the label is CIFAR-10's label.}
\label{figure:similar_compare}
 \vspace{-.6cm}
\end{figure}

\noindent\newline{\textbf{Sample-based explanations:}}
Fig. \ref{figure:similar_compare} compares our method (with equal weights) with the basic approach of calculating similarity based on the Euclidean distances of the extracted features (that is, the activation before the fully connected decition layer of the baseline model).
Qualitatively we observe that the proposed method is very successful at discovering visually similar images from the training dataset. We observe that images whose extracted features are close to the test image tend to be of the same class, but are not visually similar, whereas our proposed approach indicates the baseline model was able to discover visually similar images even if they are from different classes.

\begin{figure}[!htbp]
 \vspace{-.2cm}
\centering
\begin{subfigure}{0.49\textwidth}
  \centering
  \includegraphics[width=\linewidth]{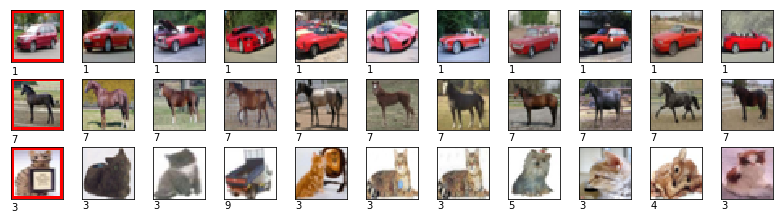}
  \vspace{-0.6cm}
\caption{Equal weights}
\end{subfigure}
\begin{subfigure}{0.49\textwidth}
  \centering
  \includegraphics[width=\linewidth]{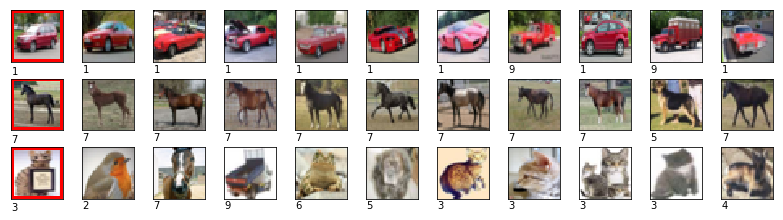}
  \vspace{-0.6cm}
  \caption{Weights: 2, 1, 0, 0}
\end{subfigure}
\begin{subfigure}{0.49\textwidth}
  \centering
  \includegraphics[width=\linewidth]{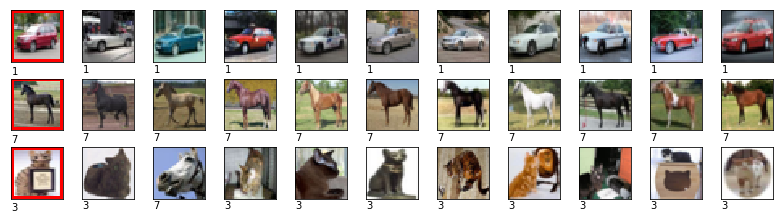}
  \vspace{-0.6cm}
\caption{Weights: 0, 0, 1, 2}
\end{subfigure}
  \vspace{-0.3cm}
\caption{Comparison of similar images with different weights for the layers of $\text{activation\_18}$, $\text{activation\_36}$, $\text{activation\_54}$. (b) focuses primarily on basic shape and color patterns and often find images similar at first sight. (c) captures the similar images with correct labels, but ignores low-level similarity patterns. (a) captures the benefits of both (b) and (c).}
\label{figure:similar_weights}
\end{figure}

\noindent\newline{\textbf{Impact of the weights of layers:}}
It is worth noting that the proposed method for ranking similar examples relies on the user to set the weight given to different layers. Fig. \ref{figure:similar_weights} shows the effect of changing the weights. We observe that focusing on lower-layer tends to assign similarity based on colors and shape, whole higher-layer tends to assign similarity based on output classes while ignoring visual similarity. Equal weights seem as a reasonable tradeoff to capture both.

\begin{figure*}[!htbp]
 \vspace{-.2cm}
\centering
  \centering
  \includegraphics[width=0.77\linewidth]{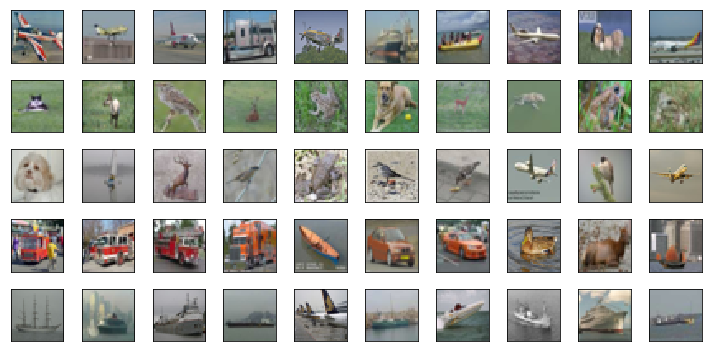}
\vspace{-0.5cm}
  \caption{Low-level concepts are based on the first selected activation. Each row of images represent a single concept.  Low-level concepts mostly focus on basic color and shape patterns, and often group different classes together. E.g. a cat and a frog images are from the same low-level concept, as they both have similar outlined shape, and similar grayish/greenish colors.}
\label{figure:concepts_low}
\end{figure*}

\begin{figure*}[!htbp]
 \vspace{-.4cm}
\centering
\includegraphics[width=0.75\linewidth]{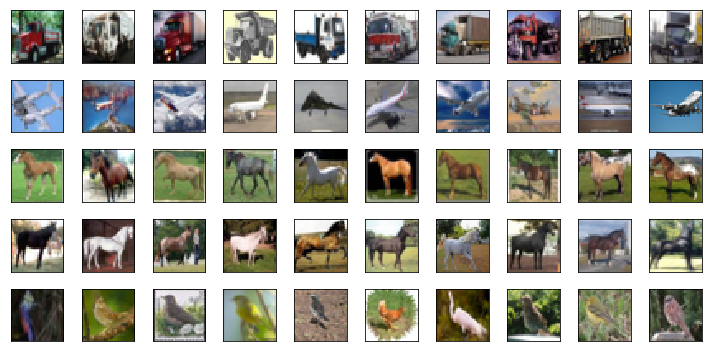}
\vspace{-0.5cm}
\caption{High-level concepts are based on the last selected activation (``$\text{activation\_54}$"). High-level concepts focus on the specifics of the predicted label and the concepts are often from the same label. We sometimes observe split of the categories beyond the labels, e.g. although the `horse' class includes both body and head images, the concepts are purely body images facing one direction.}
\label{figure:concepts_high}
\end{figure*}

\begin{figure*}[!htbp]
 \vspace{-.2cm}
\centering
  \includegraphics[width=0.8\linewidth]{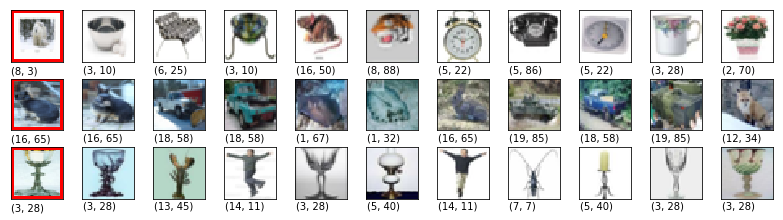}
\vspace{-0.3cm}
\caption{Similar images from CIFAR-100 using equal weights.  For each image, the labels are CIFAR-100's (coarse label, fine label).  We often observe very high similarity both in the details and the content of the images.}
\label{figure:cifar100_similar}
\end{figure*}

\begin{figure*}[!htbp]
 \vspace{-.3cm}
\centering
  \includegraphics[width=0.8\linewidth]{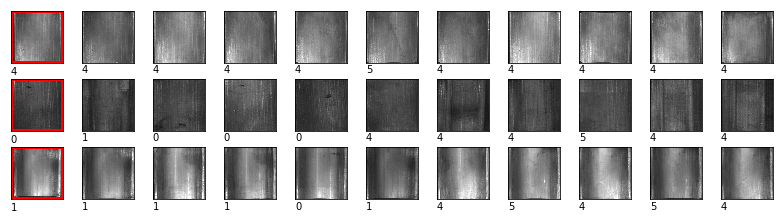}
\vspace{-0.4cm}
\caption{Similar images from the magnetic tile defect dataset. Fine labels corresponding to types defects are shown, with label `4' corresponding to NOT DEFECTIVE.}
\label{figure:mt_similar}
 \vspace{-.5cm}
\end{figure*}
\noindent\newline{\textbf{Learned concepts:}}
Figs. \ref{figure:concepts_low} and \ref{figure:concepts_high} show training images that represent the concepts learned by the baseline model, namely the images whose embeddings are nearest to the cluster centroids. We observe that the low level concepts tend to be about the color palette, whereas the high level concepts contain the categorization information that is finer than the class labels themselves.
\vspace{-0.3cm}
\subsection{CIFAR100 dataset}
\vspace{-0.3cm}

\noindent\newline{\textbf{Setup:}}
On the CIFAR-100 dataset, we fine-tune a ResNet-V2 model pre-trained on ImageNet with the $20$ \emph{coarse} labels of CIFAR-100. The resulting baseline model has an accuracy of $0.59$. Then we apply our activation clustering surrogate model, and we obtain an accuracy of $0.51$ and fidelity $0.61$.
\noindent\newline{\textbf{Sample-based explanations:}}
We observe that our proposed approach is able to capture image composition and identify images from different classes as similar.
\noindent\newline{\textbf{User study on similar examples:}}
To validate that humans would agree with our proposed approach's notion of ``similar images", we conduct a user study with 18 users and a total of 316 trials. During each trial the user is presented with one random test image and 12 training images, 3 images each from different approaches, and then asked to select the training image they consider most similar to the test image. As shown in Table \ref{table:user_study_cifar100}, our method yields the most similar images that the users agree on. It even outperforms the case when the images were chosen from the same fine label, which corresponds to extra information that does not exist for training. Note that our method is applied on a baseline model with an accuracy of 0.59, thus, it find very similar examples to the test one even when it is mis-classified.

 \vspace{-.3cm}
\begin{table}[!htbp]
 \vspace{-.3cm}
\centering
\caption{User study results for similar images.}
\begin{tabular}{|C{5.5 cm}|C{1.5 cm}|} 
\hline
\textbf{Sampled randomly from} & \textbf{Count} \\ \hline
Ten most similar images based on our proposed method &  130 \\ \hline
Ten nearest neighbors based on extracted features & 56 \\ \hline
All training images, filtering the same coarse label with prediction & 13 \\ \hline
All training images, filtering the same fine label with test image &  117 \\ \hline
\end{tabular}
\label{table:user_study_cifar100}
\end{table}



\vspace{-0.4cm}
\subsection{Magnetic tile defect dataset}
\vspace{-0.3cm}

\noindent\newline{\textbf{Setup:}}
The real-world \href{https://github.com/abin24/Magnetic-tile-defect-datasets.}{magnetic tile defect dataset} consists of 1344 images of magnetic tiles with 6 labels consisting of no defect and 5 types of defects. We train a baseline ResNet model to an accuracy of 0.94 on the coarse binary label of DEFECTIVE or NOT DEFECTIVE, and trained an activation clustering model on it.
\noindent\newline{\textbf{User study on improving trust:}}
We hypothesize that humans trust a model's predictions more if the model provides similar training examples as supporting evidence.  We conduct a user study with 16 users in 419 trials to measure the effect of human trust when the supporting evidence examples come from different sources.
In each trial, the user is presented with a test image, the model's prediction of DEFECTIVE or NOT DEFECTIVE, and three supporting evidence images (whose labels are not shown to the user) which are sampled according to different approaches. Each user is asked to score their trust in the model's prediction on a scale of 1 to 5. Table \ref{table:user_study_metal} shows that our proposed method is very effective in improving users' trust into the model. It even outperforms the case of using the same fine labels with the test image (which is an extra information not available during training). As the baseline DNN is not always accurate, trustworthiness of the proposed method can be further increased by filtering only the support examples that come from the same label with the predicted one.

\vspace{-0.3cm}
\begin{table}[!htbp]
\centering
\caption{User study results for magnetic tile defect detection.}
\begin{tabular}{|C{5.2 cm}|C{1.9 cm}|} 
\hline
\textbf{Sampled randomly from} & \textbf{Score} (Mean $\pm$ 95$\%$ CI) \\ \hline
Ten most similar images based on the proposed method, filtering the same coarse label with prediction &  3.91 $\pm$ 0.24 \\ \hline
Ten most similar images based on the proposed method & 3.67 $\pm$ 0.23 \\ \hline
All training images, filtering the same coarse label with prediction & 3.01 $\pm$ 0.22 \\ \hline
All training images &  2.59 $\pm$ 0.26 \\ \hline
All training images, filtering the same fine label as the test image (which was not known to the baseline model) &  2.95 $\pm$ 0.27 \\ \hline
\end{tabular}
\label{table:user_study_metal}
\end{table}

\vspace{-0.3cm}
\section{Conclusions}
\vspace{-0.2cm}

We propose a novel method to explain a given trained DNN based on unsupervised clustering of activations. Our method enables systematic exploration of similarities of samples based on low-level vs. high-level information content. Besides, it provides insights on the concepts learned by the DNN. Lastly, we demonstrate how it can be used to improve the trust of users in trained DNNs.

\newpage
\bibliography{references}

\begin{thebibliography}{24}
\providecommand{\natexlab}[1]{#1}
\providecommand{\url}[1]{\texttt{#1}}
\expandafter\ifx\csname urlstyle\endcsname\relax
  \providecommand{\doi}[1]{doi: #1}\else
  \providecommand{\doi}{doi: \begingroup \urlstyle{rm}\Url}\fi

\bibitem[Arik \& Pfister(2019)Arik and Pfister]{attention_prototypical}
Arik, S.~{\"{O}}. and Pfister, T.
\newblock Attention-based prototypical learning towards interpretable,
  confident and robust deep neural networks.
\newblock \emph{arXiv:1902.06292}, 2019.

\bibitem[Arrieta et~al.(2019)Arrieta, Díaz-Rodríguez, Ser, Bennetot, Tabik,
  Barbado, García, Gil-López, Molina, Benjamins, Chatila, and
  Herrera]{alej2019explainable}
Arrieta, A.~B., Díaz-Rodríguez, N., Ser, J.~D., Bennetot, A., Tabik, S.,
  Barbado, A., García, S., Gil-López, S., Molina, D., Benjamins, R., Chatila,
  R., and Herrera, F.
\newblock Explainable artificial intelligence (xai): Concepts, taxonomies,
  opportunities and challenges toward responsible ai.
\newblock 2019.

\bibitem[Caron et~al.(2018)Caron, Bojanowski, Joulin, and Douze]{deep_cluster1}
Caron, M., Bojanowski, P., Joulin, A., and Douze, M.
\newblock Deep clustering for unsupervised learning of visual features.
\newblock \emph{arxiv:1807.05520}, 2018.

\bibitem[Caron et~al.(2019)Caron, Bojanowski, Mairal, and
  Joulin]{deep_cluster2}
Caron, M., Bojanowski, P., Mairal, J., and Joulin, A.
\newblock Leveraging large-scale uncurated data for unsupervised pre-training
  of visual features.
\newblock \emph{arXiv:1905.01278}, 2019.

\bibitem[Erhan et~al.(2009)Erhan, Bengio, Courville, and Vincent]{visualizeHF}
Erhan, D., Bengio, Y., Courville, A., and Vincent, P.
\newblock Visualizing higher-layer features of a deep network.
\newblock In \emph{Technical report}, 2009.

\bibitem[Frosst \& Hinton(2017)Frosst and Hinton]{frosst2017distilling}
Frosst, N. and Hinton, G.
\newblock Distilling a neural network into a soft decision tree.
\newblock \emph{arXiv:1711.09784}, 2017.

\bibitem[Ghorbani et~al.(2019)Ghorbani, Wexler, Zou, and
  Kim]{ghorbani2019automatic}
Ghorbani, A., Wexler, J., Zou, J., and Kim, B.
\newblock Towards automatic concept-based explanations.
\newblock \emph{arXiv:1902.03129}, 2019.

\bibitem[Kahneman(2011)]{kahneman2011thinking}
Kahneman, D.
\newblock \emph{Thinking, fast and slow}.
\newblock New York, 2011.

\bibitem[Kim et~al.(2016)Kim, Khanna, and Koyejo]{kim2016examples}
Kim, B., Khanna, R., and Koyejo, O.~O.
\newblock Examples are not enough, learn to criticize! criticism for
  interpretability.
\newblock In \emph{Advances in Neural Information Processing Systems}, 2016.

\bibitem[Kim et~al.(2017)Kim, Wattenberg, Gilmer, Cai, Wexler, Viegas, and
  Sayres]{kim2017interpretability}
Kim, B., Wattenberg, M., Gilmer, J., Cai, C., Wexler, J., Viegas, F., and
  Sayres, R.
\newblock Interpretability beyond feature attribution: Quantitative testing
  with concept activation vectors.
\newblock \emph{arXiv:1711.11279}, 2017.

\bibitem[Koh \& Liang(2017)Koh and Liang]{influence_functions}
Koh, P.~W. and Liang, P.
\newblock Understanding black-box predictions via influence functions.
\newblock \emph{arXiv:1703.04730}, 2017.

\bibitem[Li et~al.(2017)Li, Liu, Chen, and Rudin]{li_protototypes}
Li, O., Liu, H., Chen, C., and Rudin, C.
\newblock Deep learning for case-based reasoning through prototypes: {A} neural
  network that explains its predictions.
\newblock \emph{arXiv:1710.04806}, 2017.

\bibitem[Lundberg \& Lee(2017)Lundberg and Lee]{lundberg2017unified}
Lundberg, S.~M. and Lee, S.-I.
\newblock A unified approach to interpreting model predictions.
\newblock In \emph{Advances in Neural Information Processing Systems}, pp.\
  4765--4774, 2017.

\bibitem[Ribeiro et~al.(2016)Ribeiro, Singh, and Guestrin]{lime}
Ribeiro, M.~T., Singh, S., and Guestrin, C.
\newblock "why should {I} trust you?": Explaining the predictions of any
  classifier.
\newblock \emph{arXiv:1602.04938}, 2016.

\bibitem[{Rudin}(2018)]{rudin_explainableAI}
{Rudin}, C.
\newblock {Please Stop Explaining Black Box Models for High Stakes Decisions}.
\newblock \emph{arXiv:1811.10154}, 2018.

\bibitem[Shrikumar et~al.(2017)Shrikumar, Greenside, and Kundaje]{deeplift}
Shrikumar, A., Greenside, P., and Kundaje, A.
\newblock Learning important features through propagating activation
  differences.
\newblock In \emph{International Conference on Machine Learning-Volume}, pp.\
  3145--3153, 2017.

\bibitem[Simonyan et~al.(2013)Simonyan, Vedaldi, and Zisserman]{visualizeconv2}
Simonyan, K., Vedaldi, A., and Zisserman, A.
\newblock Deep inside convolutional networks: Visualising image classification
  models and saliency maps.
\newblock \emph{arXiv:1312.6034}, 2013.

\bibitem[Tan et~al.(2018)Tan, Caruana, Hooker, Koch, and
  Gordo]{tan2018learning}
Tan, S., Caruana, R., Hooker, G., Koch, P., and Gordo, A.
\newblock Learning global additive explanations for neural nets using model
  distillation.
\newblock \emph{arXiv:1801.08640}, 2018.

\bibitem[Xie et~al.(2015)Xie, Girshick, and Farhadi]{xie2015unsupervised}
Xie, J., Girshick, R., and Farhadi, A.
\newblock Unsupervised deep embedding for clustering analysis.
\newblock \emph{arXiv:1511.06335}, 2015.

\bibitem[Yeh et~al.(2018)Yeh, Kim, Yen, and Ravikumar]{yeh2018representer}
Yeh, C.-K., Kim, J.~S., Yen, I. E.~H., and Ravikumar, P.
\newblock Representer point selection for explaining deep neural networks.
\newblock \emph{arXiv:1811.09720}, 2018.

\bibitem[Yeh et~al.(2019)Yeh, Kim, Arik, Li, Pfister, and
  Ravikumar]{yeh2019completenessaware}
Yeh, C.-K., Kim, B., Arik, S.~O., Li, C.-L., Pfister, T., and Ravikumar, P.
\newblock On completeness-aware concept-based explanations in deep neural
  networks.
\newblock \emph{arXiv:1910.07969}, 2019.

\bibitem[Yoon et~al.(2019)Yoon, Arik, and Pfister]{yoon2019rllim}
Yoon, J., Arik, S.~O., and Pfister, T.
\newblock Rl-lim: Reinforcement learning-based locally interpretable modeling.
\newblock \emph{arXiv:1909.12367}, 2019.

\bibitem[Zeiler \& Fergus(2014)Zeiler and Fergus]{cnn_filters1}
Zeiler, M.~D. and Fergus, R.
\newblock Visualizing and understanding convolutional networks.
\newblock In Fleet, D., Pajdla, T., Schiele, B., and Tuytelaars, T. (eds.),
  \emph{Computer Vision -- ECCV 2014}, pp.\  818--833, Cham, 2014. Springer
  International Publishing.

\bibitem[Zhou et~al.(2018)Zhou, Sun, Bau, and Torralba]{Bolei2018}
Zhou, B., Sun, Y., Bau, D., and Torralba, A.
\newblock Interpretable basis decomposition for visual explanation.
\newblock In Ferrari, V., Hebert, M., Sminchisescu, C., and Weiss, Y. (eds.),
  \emph{ECCV}, 2018.

\end{thebibliography}
\bibliographystyle{icml2020}

\newpage

\onecolumn

\appendix

\section{User studies}

\begin{figure}[!ht]
\centering
  \includegraphics[width=0.7\linewidth]{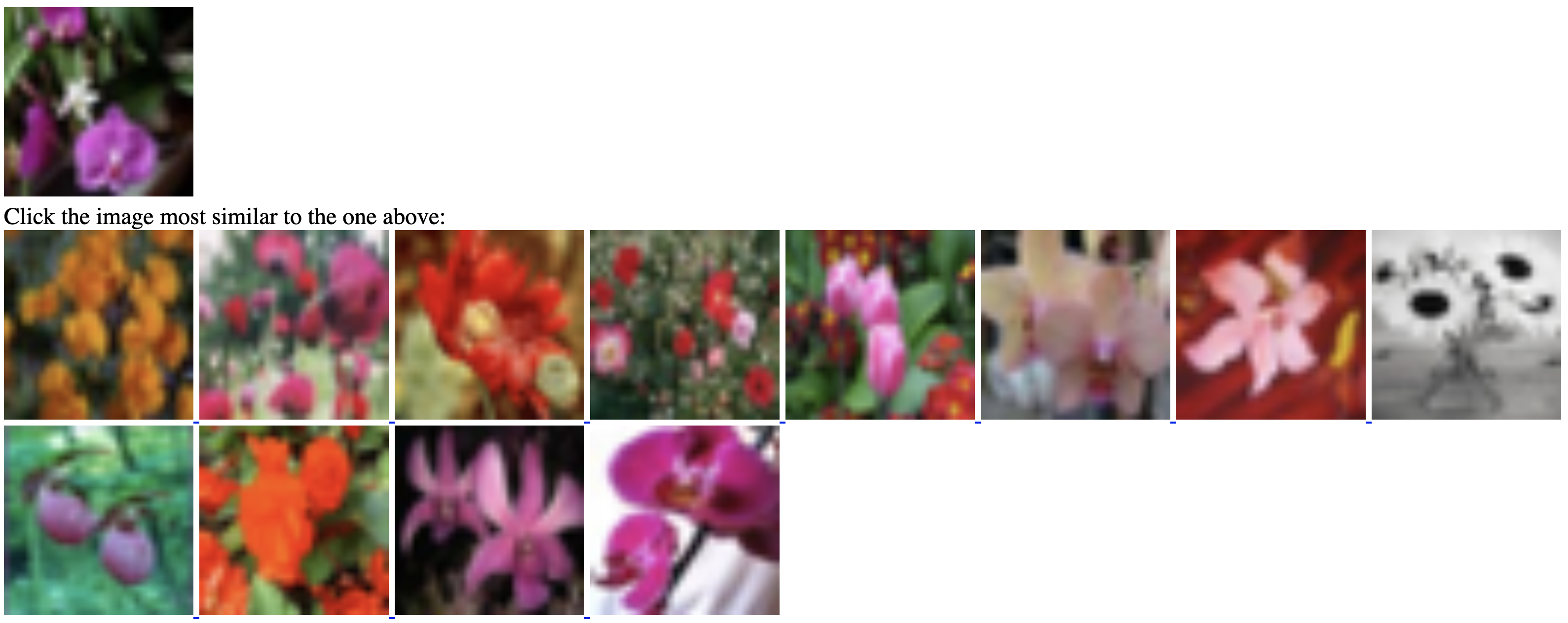}
\caption{CIFAR-100 similar image user study.}
\label{figure:cifar100_user_study}
\end{figure}

\begin{figure}[!ht]
\centering
  \includegraphics[width=0.7\linewidth]{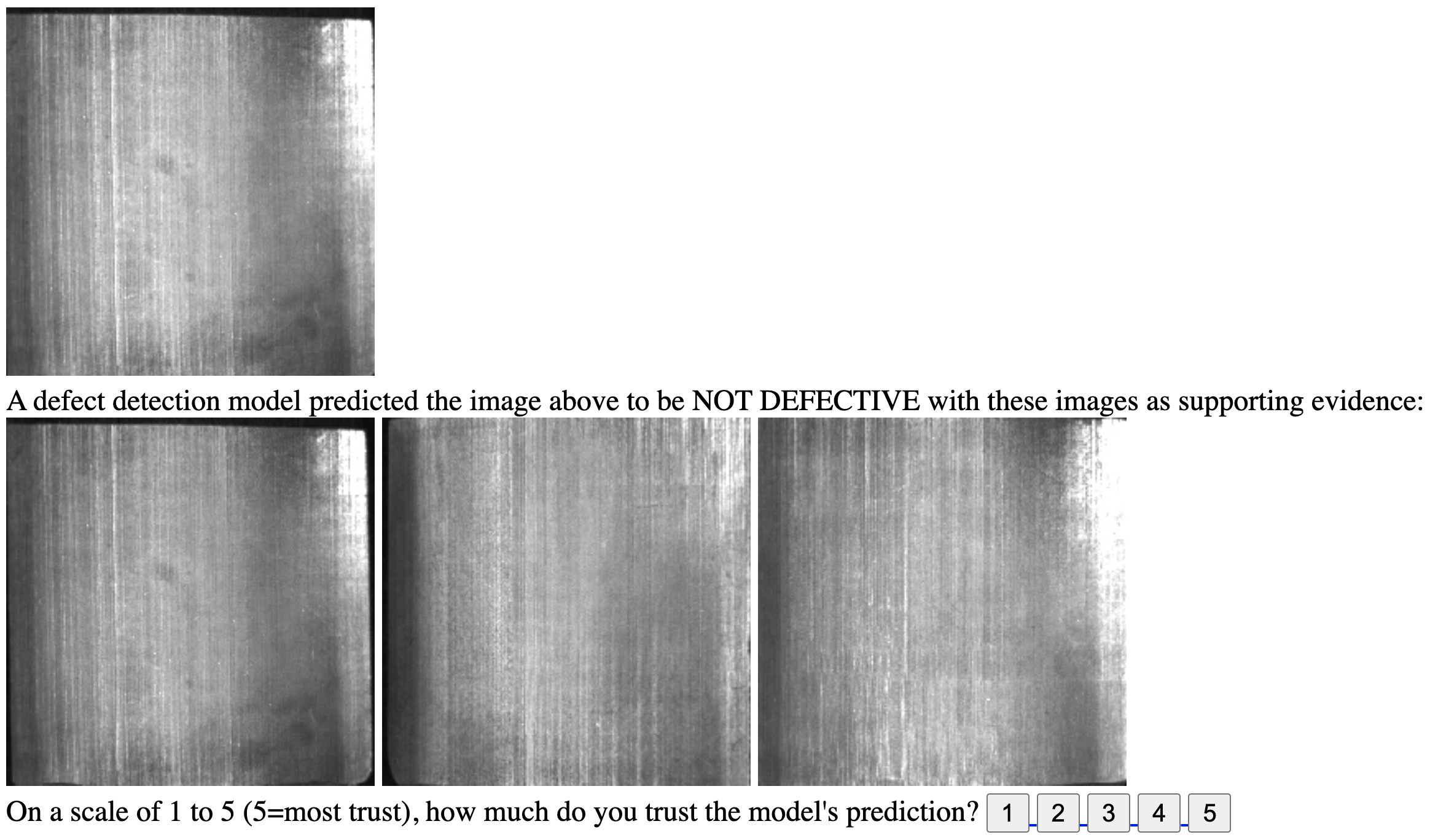}
\caption{Magnetic tiles trust user study.}
\label{figure:mt_user_study}
\end{figure}

Figs. \ref{figure:cifar100_user_study} and \ref{figure:mt_user_study} visualize the interfaces of the user studies.

\newpage
\section{More examples of similar images}

\begin{figure*}[!ht]
 \vspace{-.2cm}
\centering
  \includegraphics[width=\linewidth]{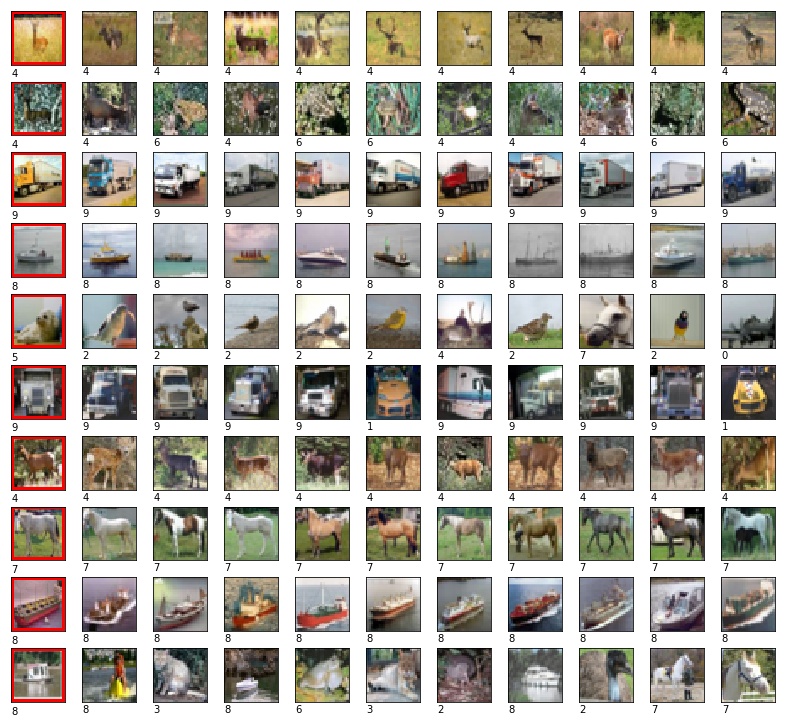}
\vspace{-0.3cm}
\caption{Similar images from CIFAR-10 using equal weights. The left most column has the test images with red border, and ten most similar images are shown.  For each image, the label is CIFAR-10's label.}
\label{figure:appendix_cifar10_similar}
 \vspace{-.4cm}
\end{figure*}

\begin{figure*}[!ht]
 \vspace{-.2cm}
\centering
  \includegraphics[width=\linewidth]{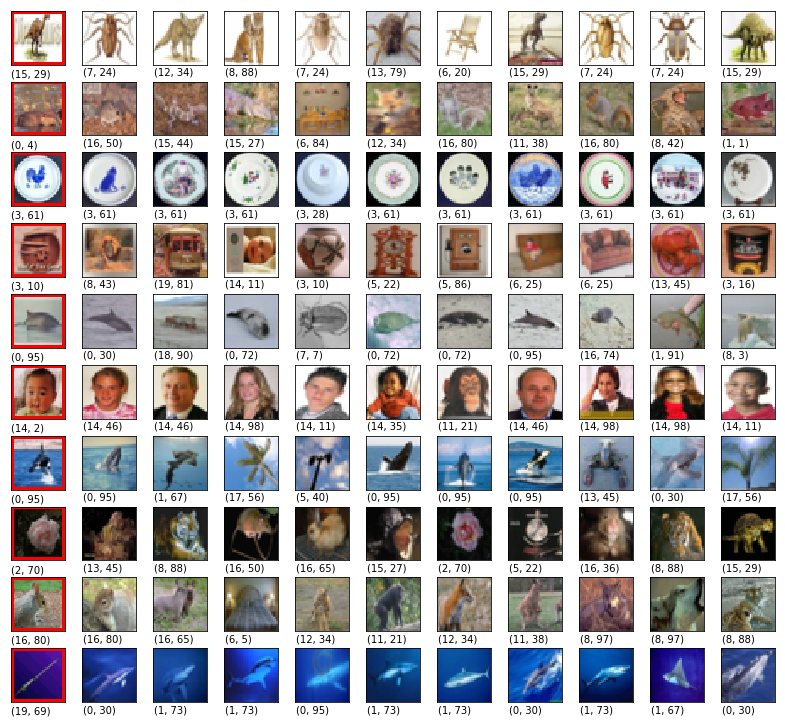}
\vspace{-0.3cm}
\caption{Similar images from CIFAR-100 using equal weights. The left most column has the test images with red border, and ten most similar images are shown.  For each image, the labels are CIFAR-100's (coarse label, fine label).}
\label{figure:appendix_cifar100_similar}
 \vspace{-.4cm}
\end{figure*}

\begin{figure*}[!htbp]
 \vspace{-.2cm}
\centering
\begin{subfigure}{\textwidth}
  \centering
  \includegraphics[width=\linewidth]{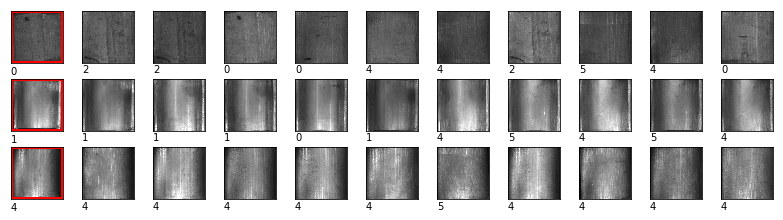}
  \vspace{-0.3cm}
\caption{Equal weights}
\end{subfigure}
\begin{subfigure}{\textwidth}
  \centering
  \includegraphics[width=\linewidth]{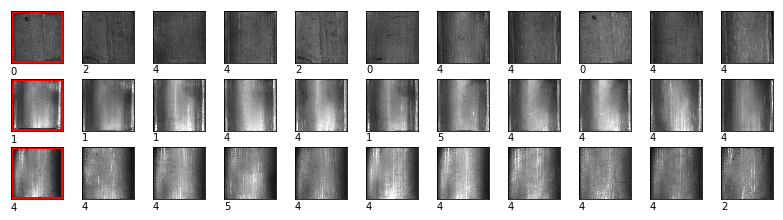}
  \vspace{-0.3cm}
  \caption{Weights: 2, 1, 0, 0}
\end{subfigure}
\begin{subfigure}{\textwidth}
  \centering
  \includegraphics[width=\linewidth]{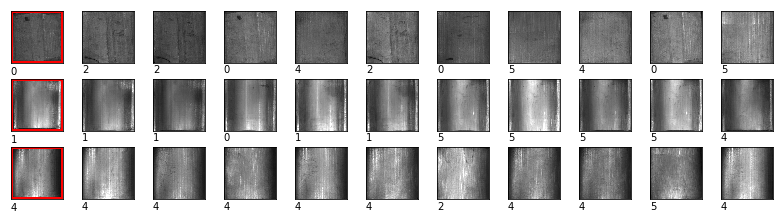}
  \vspace{-0.3cm}
\caption{Weights: 0, 0, 1, 2}
\end{subfigure}
\caption{Comparison of similar images with different weights for magnetic tile defect dataset.}
\label{figure:appendix_similar_weights_mt}
\end{figure*}

Figs. \ref{figure:appendix_cifar10_similar}, \ref{figure:appendix_cifar100_similar} and \ref{figure:appendix_similar_weights_mt} show more examples from CIFAR-10, CIFAR-100 and magnetic tile defect datasets respectively.

\clearpage
\newpage
\section{More examples of discovered concepts}

\begin{figure*}[!htbp]
 \vspace{-.2cm}
\centering
  \centering
  \includegraphics[width=0.7\linewidth]{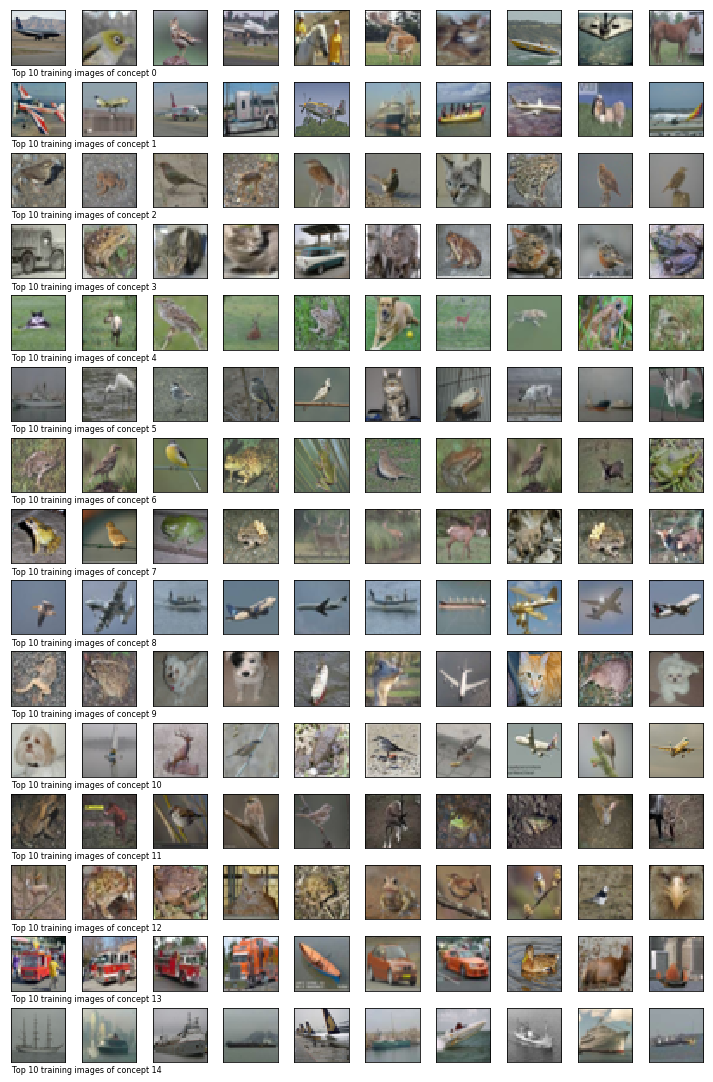}
\vspace{-0.4cm}
  \caption{Low-level concepts are based on the first selected activation. Low-level concepts mostly focus on basic color and shape patterns, and often group different classes together. E.g. a cat and a frog images are from the same low-level concept, as they both have similar outlined shape, and similar grayish/greenish colors.}
\label{figure:concepts_low_full}
\end{figure*}

\begin{figure*}[!htbp]
 \vspace{-.4cm}
\centering
\includegraphics[width=0.7\linewidth]{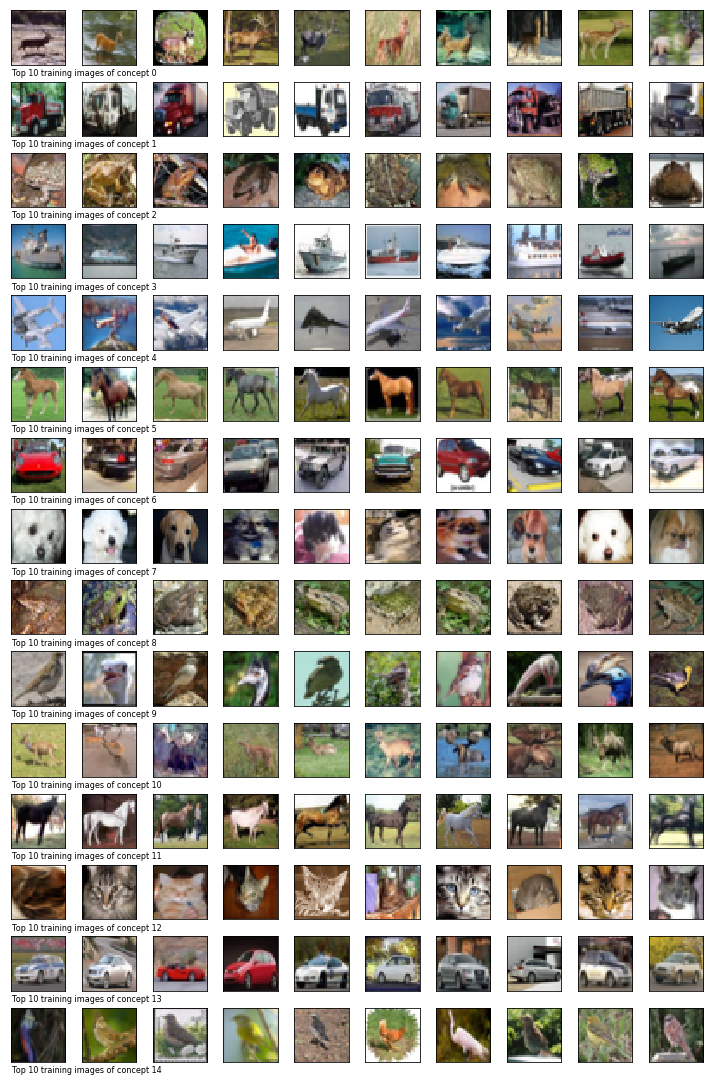}
\vspace{-0.3cm}
\caption{High-level concepts are based on the last selected activation (``$\text{activation\_54}$"). High-level concepts focus on the specifics of the predicted label and the concepts are often from the same label. We sometimes observe split of the categories beyond the labels, e.g. although the `horse' class includes both horse body and head images, while the corresponding are purely body images facing one direction.}
\label{figure:concepts_high_full}
\end{figure*}

\end{document}